\documentclass{article}

\usepackage{arxiv}
\usepackage{float}
\usepackage[T1]{fontenc}    % use 8-bit T1 fonts
\usepackage{hyperref}       % hyperlinks
\usepackage{url}            % simple URL typesetting
\usepackage{graphicx}
\graphicspath{ {./images/} }
\usepackage{multirow}
\usepackage{xcolor}
\usepackage{array}
\newcolumntype{P}[1]{>{\centering\arraybackslash}p{#1}}
\definecolor{OliveGreen}{rgb}{0,0.6,0}
\definecolor{OliveBlue}{rgb}{0,0,0.6}
\definecolor{OliveRed}{rgb}{0.6,0,0}
\definecolor{AllOlive}{rgb}{0.6,0.6,0.6}
\definecolor{pink}{rgb}{0.8,0.6,0.2}
\definecolor{yellow}{rgb}{1,1,0}
\definecolor{purple}{rgb}{0.8,0.4,1}
\title{Techniques to Improve Q\&A Accuracy with Transformer-based models on Large Complex Documents }

\date{} 					% Or removing it

\author {\href{https://orcid.org/0000-0003-4939-5034}{\includegraphics{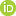}\hspace{1mm}Chejui Liao}\\
Synechron Innovation Lab\\
\texttt{jerry.liao@synechron.com}\\
\And
\href{https://orcid.org/0000-0002-7895-6681}{\includegraphics{images/ORCIDiD_icon16x16.png}\hspace{1mm}Tabish Maniar}\\
Synechron Innovation Lab\\
\texttt{tabish.maniar1@synechron.com}\\
\And
\href{https://orcid.org/0000-0002-4530-9533}{\includegraphics{images/ORCIDiD_icon16x16.png}\hspace{1mm}Sravanajyothi N}\\
Synechron Innovation Lab\\
\texttt{sravanajyothi.n@synechron.com}\\
\And
\href{https://orcid.org/0000-0002-9064-3362}{\includegraphics{images/ORCIDiD_icon16x16.png}\hspace{1mm}Anantha Sharma}\\
Synechron Innovation Lab\\
\texttt{anantha.sharma@synechron.com}\\
}

\hypersetup{
pdftitle={Text Processing Techniques to Improve Accuracy in a BERT based Q\&A System},
pdfsubject={cs.LG, stat.ML},
pdfauthor={Chejui Liao, Tabish Maniar, Sravanajyothi N, Anantha Sharma},
pdfkeywords={BERT, document processing, QnA model},
}

\begin{document}
\maketitle
    
\begin{abstract}
	This paper discusses the effectiveness of various text processing techniques, their combinations, and encodings to achieve a reduction of complexity and size in a given text corpus. The simplified text corpus is sent to BERT (or similar transformer based models) for question and answering and can produce more relevant responses to user queries. This paper takes a scientific approach to determine the benefits and effectiveness of various techniques and concludes a best-fit combination that produces a statistically significant improvement in accuracy. 
\end{abstract}

% keywords can be removed

\section{Introduction}
\label{sec:intro}
In today's world, BERT \cite{devlin2018bert} is one of the most popular models used to build question and answering systems. BERT generally needs additional training (with large relevant text corpus) to maintain relevance when presented with large complex documents (such as regulations, federal or institutional policies, and domain-specific documents). This shortcoming  becomes clear when dealing with a complex sentence structure and can be alleviated by using innovative text pre-processing techniques on the text corpus and fine-tuning the model.

Why is this needed: 
\begin{itemize}
    \item Fine-tuned BERT QnA model can handle sequence length of 384 tokens at once. When the input context has more has 384 tokens, context gets divided into multiple chunks (length of each chunk can be set by user) and this is popularly known as “sliding window” approach.  
    \item Each chunk is then processed along with question by BERT model giving answer text and probability as output, final answer text will be considered based on highest probability criteria.  
    \item We observed that sliding window approach fails to provide correct answers when document size was over 1000 tokens, as BERT model was trained on data with smaller contexts.  
\end{itemize}

This led us to explore possibilities of reducing complexity of input text and finding most relevant text from large context before passing on to BERT for inference. This paper explores the possibility of using text processing techniques to reduce the complexity of text sending to BERT and hence improve BERT’s accuracy on answering questions for complex documents without lots of extra training. The limitations of each techniques are also discussed in this paper. 

\section{Methodology}
To improve the performance on BERT, we implemented a couple of techniques, such as Definition Tokenization, Dependency Tokenization, Paragraph Splitting, Relevant Paragraph Ranking, and BERT Fine-tuning. The entire flow for text processing and BERT question and answering system we tried to propose is as below:

\begin{figure}[ht]
  \includegraphics{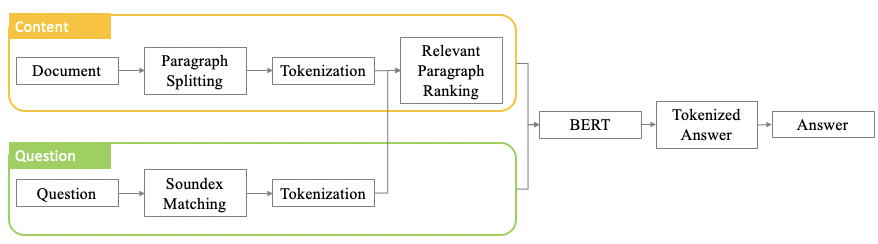}
  \caption{Proposed Text Processing Flow for BERT QnA}
\end{figure}

\subsection{Definition Tokenization}
One of our tokenization strategies is Definition Tokenization. We replace the specific terms used in the document with a single token. First, we identify definition sentences mentioned in the document via keyword search such as "mean" and "define". For example, "Common ownership means a relationship between two companies" is a definition sentence. Second, we identify the subject (including the noun and its modifiers) in the definition sentences through Stanford CoreNLP's dependency annotation \cite{manning2014stanford}, "Common ownership" in this case, and tokenize the subject as XnXn (where n is a number, X can be any capitalized characters). The result of tokenization is as below: 

\begin{center}
\begin{tabular}{ |c|c| } 
\hline
Original & \textbf{Common ownership} means a relationship between two companies. \\
\hline
Definition Tokenization & \textbf{X1X1} means a relationship between two companies. \\
\hline
\end{tabular}
\end{center}

LexNLP \cite{bommarito2018lexnlp} is also used to help us tokenize compound nouns outside of the definition sentences. LexNLP is trained on financial domains and is able to recognize some specific terms in the domain. The result is as below: 

\begin{center}
\begin{tabular}{ |c|c| } 
\hline
Original & \textbf{Financial Institution} needs to submit a suspicious activity report. \\
\hline
Definition Tokenization & \textbf{X1X2} needs to submit a suspicious activity report. \\
\hline
\end{tabular}
\end{center}

Finally, we replace all the definition subject with correspondent tokens over the document.  

With this approach, we are able to reduce a couple of words into one token, and thus reduce the amount of text in the sentences. 

The limitations of this approach include the following: 
\begin{itemize}
	\item Variant amount of tokenization: highly dependent on how often the definition terms are used across the document and how many terms are defined 
	\item Customization of the definition keywords for different documents: each document may have different keyword for identifying a definition sentence 
\end{itemize}

% See Section \ref{sec:headings}.

\subsection{Dependency Tokenization}
\label{sec:approach}
The other tokenization strategy we use is Dependency Tokenization that uses Stanford CoreNLP's dependency annotation to tokenize words. We first identify the verb in a sentence. Then we group the subjects and modifiers and tokenize them. We also do the same for objects. Below is the result: 

\begin{center}
\begin{tabular}{ |c|c| } 
\hline
Original & \textbf{Bank and insurance company} need to submit \textbf{a suspicious activity report}. \\
\hline
Definition Tokenization & \textbf{X1X3} need to submit \textbf{X1X4}. \\
\hline
\end{tabular}
\end{center}

In this example, “bank" and “company" are subjects of “need", and “insurance" is the modifier of “company", so we tokenize them all together. “report" is the object in this case, “a", “suspicious", “activity" are its modifiers. Complex sentences usually involve a lot of subjects, objects, and modifiers. With this approach, we can simplify such sentences to a great extent.

The limitations of this approach include the following: 
\begin{itemize}
	\item Highly dependent on the accuracy of NLP dependency parse: in very complicated sentences, the dependency results may not be correct and end up tokenizing wrong words, distorting the structure of the sentences  

    \item Computationally expensive: extracting dependency from each sentence requires a lot of computation 
\end{itemize}

\subsection{Paragraph Splitting}

The other text processing we do is to split the document into paragraphs and send the most relevant paragraphs with a question for BERT to answer. Since the document we are dealing with contains tens of thousands of words, BERT is not able to pick up the answer in a sea. Given the hierarchy of the regulation documents we deal with, we split the document into small paragraphs (which contain a piece of information about a specific regulation) using regular expression and spaCy's sentence segmentation \cite{spacy2}.  

The limitations of this approach include the following:   
\begin{itemize}
    \item BERT not able to answer broader questions: after splitting into paragraphs, the information becomes more specific, and BERT cannot revert to high-level answer 

    \item Splitting strategy very subjective: developer needs to make judgement about how far the splitting should go. If the paragraph is too long, BERT will still have issue figuring out answer; however, if the paragraph is too short, you may not have enough information to match the question. The golden rule is that the split paragraph should contain complete piece of information  
\end{itemize}
\begin{center}
\begin{tabular}{|c@{}|c@{}|c@{}|}
\hline
Original & 
\begin{tabular}{c}
     5 times the amount of the \textbf{\textcolor{OliveRed}{nonvoting capital stock}}\\ \textbf{\textcolor{OliveGreen}{of the Financing Corporation} }  \\
    which is outstanding at \textbf{\textcolor{purple}{such time}};\\ 
    or the \textbf{\textcolor{OliveBlue}{amount of capital stock}} of the  \\    
    \textbf{\textcolor{pink}{Financing Corporation}} held by\\ \textbf{\textcolor{orange}{such remaining bank}} at the time of such \\
    \textbf{\textcolor{cyan}{determination}}; by \textbf{\textcolor{brown}{the amounts}} added to reserves\\ after December 31, 1985, \\
    pursuant to the requirement contained\\ in the first 2 sentences of section 1436 \\
    of this title.
\end{tabular}
& Number of Tokens :66
\\\hline
\begin{tabular}{c}
     Definition  \\
     and \\
     Dependency\\
     Tokenization
\end{tabular}
  & 
\begin{tabular}{c}
     5 times the amount of the \\ \textcolor{OliveRed}{Y1Y300} \textcolor{OliveGreen}{X1441} which is \\outstanding at \textcolor{purple}{Y1Y1416} ; or  \\
     the amount of \textcolor{OliveBlue}{Y1Y1122} of the \textcolor{pink}{Y1Y415} held by such\\ \textcolor{orange}{Y1Y1099} at the time of \textcolor{cyan}{Y1Y651} \\
     \textcolor{brown}{X1393} added to reserves after december 31 , 1985 ,\\
     pursuant to the requirement\\ contained in the first 2 sentences of section 1436 of \\this title
\end{tabular}
& Number of Tokens :54

\\\hline
\end{tabular}
\end{center}

\subsection{Relevant Paragraph Ranking}

To narrow BERT’s search space and obtain more accurate answers, we build a similarity ranking model based on Doc2Vec \cite{le2014distributed} and TF-IDF \cite{ramos2003using} to locate the most relevant paragraphs given a question.  

We compare the similarity among the paragraphs and the question and send the top relevant paragraphs to BERT for an answer. The number of top paragraphs is a hyperparameter, depending on the size of the document, the larger the document, the more relevant paragraphs needed. Doc2Vec provides us the flexibility of the words used in the questions. User does not need to provide exact same words as appeared in the documents to get the answer, since the sentences are vectorized and compared according to the context; while TF-IDF helps us identify the right paragraphs more accurately if user does provide the exact words. 

Given the questions and the documents we had, we use the weight of 50-50 between Doc2Vec and TF-IDF.  

The limitations of this approach include the following: 
\begin{itemize}
    \item The weight between Doc2Vec and TF-IDF is arbitrary: increase the weight of Doc2Vec if we want more flexibility in question’s wording, but we may lose accuracy 
    \item The number of top relevant paragraphs varies: it depends on the how specific the question is, how similar the paragraphs are in the document, and how huge the document is 
\end{itemize} 

\subsection{Soundex Encoding}

To overcome spelling mistakes in questions, we also implement Soundex Based Encodings \cite{koneru2016performance}. Soundex tries to encode the words in the form of phonetic sounds. Even if the spellings are wrong in the question, the phonetic encoding remains the same. E.g. Hello yields an encoding of H400 and Hallo also yields the encoding of H400. With TF-IDF the Soundex encoded terms are used that make it easier to find the relevant sections from the document. By default, Soundex uses encodings of length 4 (H400) but we decided to use encoding of length 6 (H40000) to accommodate more term varieties in our documents. 

\subsection{BERT Fine-tuning}

 We analyzed the performance of BERT model trained on SQUAD2.0 \cite{rajpurkar2018know} and decided to fine tune it further with data distribution similar to FDIC documents since SQUAD2.0 dataset is observed to have simple structure as compared to sentence complexity structure in FDIC documents. The ones in FDIC spanned across multiple lines hence leading to increased complexity of structure. Since BERT is trained on SQUAD dataset, model was observed performing poor on FDIC documents even for simple "who/where" kind of questions. 

Training data consists of 88 paragraphs/context along with multiple questions and answers for each paragraph in the SQUAD data format. Similarly test data consists of 50 paragraphs and validation data has got 15 paragraphs following the same format as described above for train data. 

After experimenting with different set of hyperparameters, we finalized the one that used Adam as Optimizer \cite{kingma2014adam} with learning-rate=3e-5, train-batch-size=24 with other parameters having default values since this combination was proven to be optimized in terms of model accuracy when tested.

\subsection{Experimental Result}
Due to integration difficulties in tokenization, we tested BERT question and answer system with paragraph splitting and relevant paragraph ranking on three regulation documents, Suspicious Activity Reports \footnote{https://www.fdic.gov/regulations/laws/rules/2000-6000.html}, Fair Housing \footnote{https://www.fdic.gov/regulations/laws/rules/2000-6000.html}, and Appraisals \footnote{https://www.fdic.gov/regulations/laws/rules/2000-4300.html}, from the FDIC website. 

We define two metrics to evaluate BERT performance: 
\begin{itemize}
    \item F1 score: We used the same approach as in SQUAD for evaluation. In this approach an answer was divided into tokens and confusion matrix was calculated based on the comparison of tokens between BERT’s answer and real answer.\cite{sokolova2006beyond}
        \subitem True Positives (tp): the number of tokens shared between the actual answer and the predicted answer.  
        \subitem False Positives (fp): the tokens in the prediction but not in the actual answer.  
        \subitem False Negative(fn): the tokens in actual answer and not in the predicton. 
        \subitem Precision: \[\frac{tp}{(tp+fp)}\]
        \subitem Recall: \[\frac{tp}{(tp+fn)}\]
        \subitem F1 Score: \[2*\frac{precision*recall}{(precision+recall)}\]
    \item Quality score (Q score): We defined a scoring system and manually evaluate the quality of answers using the following standards. 
        \subitem 1: Unacceptable, if BERT does not cover the complete response
        \subitem 2: Partial answered, if BERT covers the partially the complete response 
        \subitem 3: Complete answered, if BERT covers the complete response to the question 
\end{itemize} 

We compared the F1 score and quality score among BERT with entire document, BERT with manually selected paragraph, and BERT with text processing techniques. Figure 2-4 are the flows for each system.

\begin{figure}[hbt]
  \includegraphics[width=15cm]{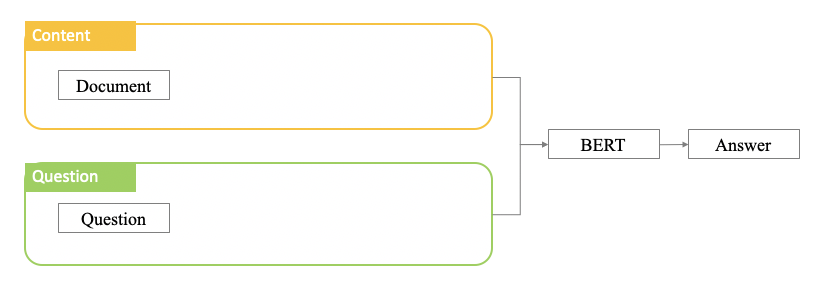}
  \caption{BERT with entire document}
\end{figure}

\begin{figure}[hbt]
  \includegraphics[width=15cm]{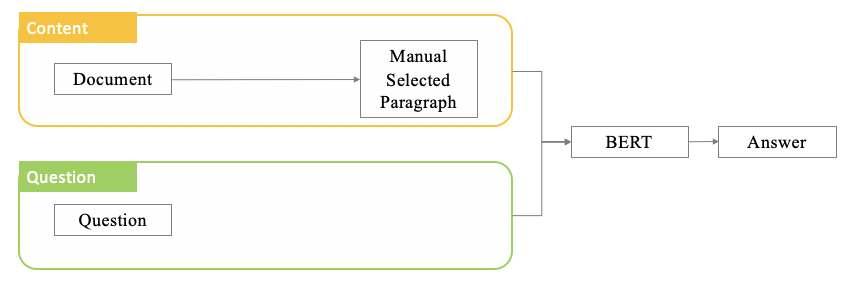}
  \caption{BERT with manually selected paragraph}
\end{figure}

\begin{figure}[hbt]
  \includegraphics[width=15cm]{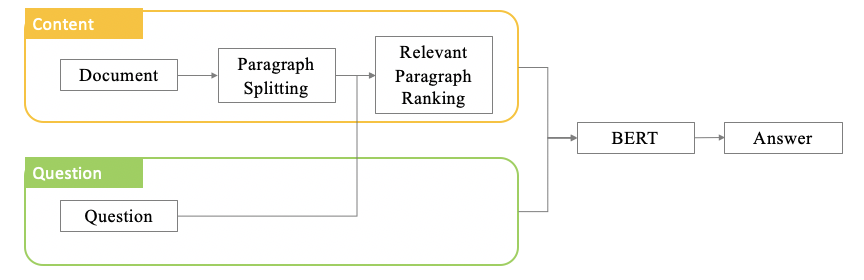}
  \caption{BERT with Text Processing}
\end{figure}

We normalized the final score as \[\frac{(Quality score-1)}{2}\] to get a score ranging from 0 to 1 for better comparison. 

\begin{table}[htbp]
    \centering
        \begin{tabular}{|c@{}|c@{}|c@{}|c@{}|c@{}|c@{}|c@{}|c@{}|c@{}|}
        % \begin{tabular}{|P{2.2cm}|P{1.5cm}|P{1.5cm}|P{2.7cm}@{}|P{2.7cm}@{}|P{2.7cm}@{}|}
            \hline
                \textbf{Document} &
                \begin{tabular}{c}
                     \textbf{Document} \\
                     \textbf{size} \\
                     \textbf{(words)}
                \end{tabular} & 
                 \begin{tabular}{c}
                     \textbf{Number} \\
                     \textbf{of}\\
                     \textbf{Questions}
                \end{tabular}& \multicolumn{2}{|c|}{\textbf{Entire document} } 
            & \multicolumn{2}{|c|} {\begin{tabular}{c}
                 \textbf{Manually}  \\
                 \textbf{selected} \\
                 \textbf{ paragraph}
            \end{tabular} } & 
            \multicolumn{2}{|c|}{\begin{tabular}{c}
                 \textbf{Text}  \\
                 \textbf{processing} \\
                 \textbf{techniques}
            \end{tabular} } \\
            \cline{4-9}
            & & & \textbf{F1 Score } & \textbf{Q Score } & \textbf{F1 Score } & \textbf{Q Score } & \textbf{F1 Score } & \textbf{Q Score } \\
            % & & &      \begin{tabular}{|l|r|}
            %                 \hline
            %                  Qscore &  F1 \\ 
            %             \end{tabular}
            % & & \\
            \hline
            \begin{tabular}{c}
                 Suspicious Activity  \\
                 Report
            \end{tabular}
            & 1420  & 27 
            & 9.7\% & 3.7\% & 58.7\% & 79.5\% & 56.6 \% & 74\%  \\
            \hline
            Fair Housing & 1780  & 17 
            & 8.2 \% & 0\% & 44.6\% & 67.5\% & 41.3\% & 61.5\% \\
            \hline
            Appraisals & 5367  & 14 
            & 6.42 \% & 0\% & 62.1\% & 61\% & 51.7\% & 50\% \\
            \hline
        \end{tabular}
    \caption{Test Results}
    \label{tab:my_label}
\end{table}

\begin{figure}[H]
  \includegraphics[width=15cm]{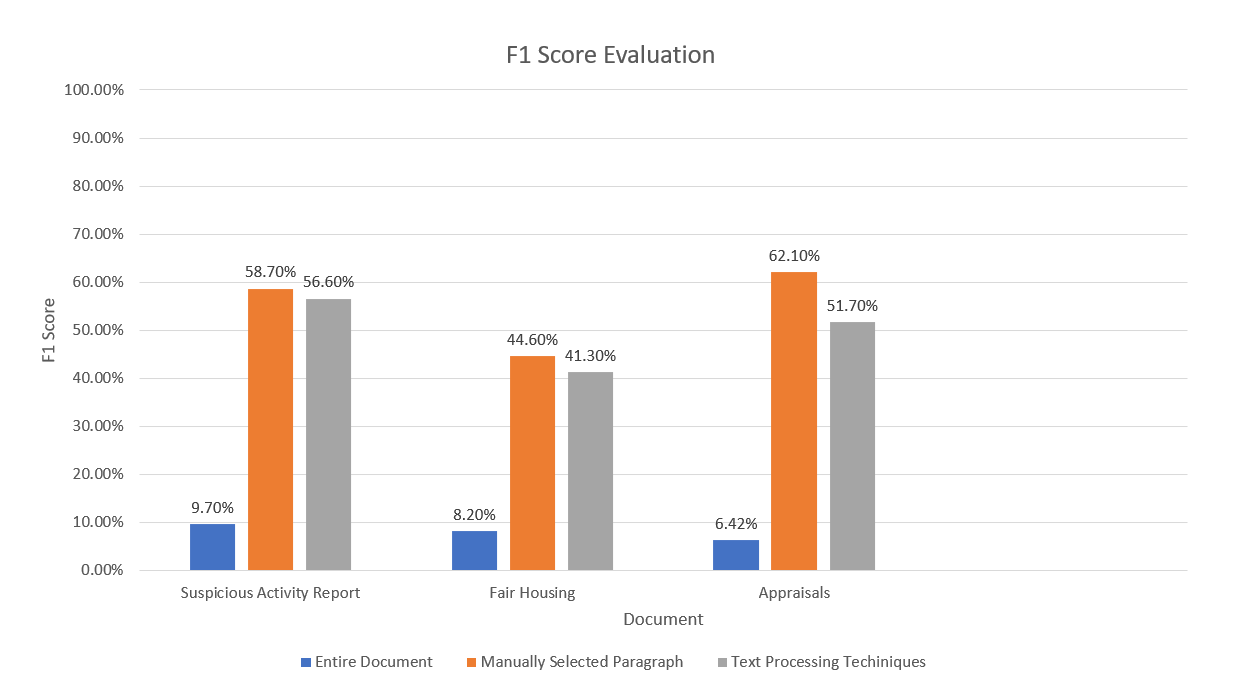}
  \caption{F1 Score comparison}
\end{figure}
\section{Conclusion}
\label{sec:conclusion}

\begin{figure}[H]
  \includegraphics[width=15cm]{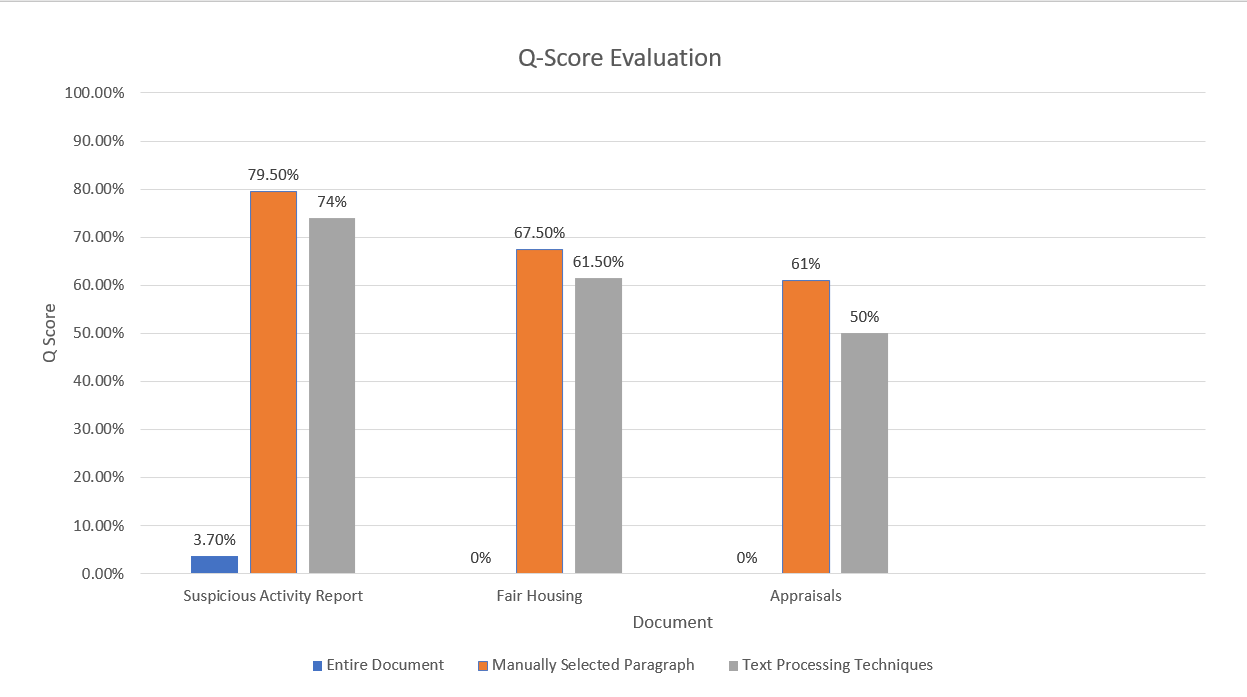}
  \caption{Q-Score comparison}
\end{figure}

With text processing techniques such as paragraph splitting and relevant paragraph ranking, we can narrow down the content size and boost BERT accuracy on large documents by 30-50\% in terms of F1 score. The upper bound of our BERT performance is BERT with manually selected paragraph since we directly provide the most relevant content. Compared to the upper bound, BERT with text processing techniques only sacrifice about 5-11\% accuracy on F1 score.

Although Tokenization techniques also look promising for simplifying complex documents, during our development of the algorithm, we ran into the following issues to integrate this solution with our BERT QnA system: 

\begin{itemize}
    \item The handling of duplicate tokens: We should use the same tokens for the same phrases over the document. 

    \item The matching of question and tokens: Dependency tokenization tokenized phrases in a more varied fashion (Adjective and Adverb are also tokenized with the noun), making it difficult to match with questions since user may not provide exact phrases as in the document. Although we tried to iterate through all the possible matching tokens, we were not able to determine which token should be taken since BERT could not provide us the absolute probability of an answer. 
\end{itemize} 

In the future, we will continue to improve our Tokenization algorithm, so it can fit into our BERT QnA system. One of the approaches to improve the tokenization algorithm is to determine the acronyms in the document and their meaning and linking the sections and their meaning in the documents\cite{saumya2020summaryclass}.We will also continue to explore more text processing techniques to improve BERT’s performance. 

\bibliographystyle{apalike}
\bibliography{bibliography.bib}
\end{document}